\documentclass[11pt]{article}
\usepackage{times}
\usepackage{authblk}
\usepackage{latexsym}
\usepackage{amsmath}
\usepackage{amssymb}
\usepackage{graphicx}
\usepackage{rotating}
\usepackage{verbatim}
\usepackage{url}
\usepackage{makecell}
\usepackage{breqn}

\usepackage{subcaption} 
\usepackage{subfiles}
\usepackage{verbatim}
\usepackage{booktabs}
\usepackage{pifont}
\usepackage[final]{neurips}
\usepackage{xcolor}
\usepackage{chngpage}
\usepackage{multirow}
\usepackage[colorlinks,allcolors=black]{hyperref}
\usepackage{cleveref}

\usepackage{etoolbox}
\makeatletter
\pretocmd{\NAT@citex}{%
  \let\NAT@hyper@\NAT@hyper@citex
  \def\NAT@postnote{#2}%
  \setcounter{NAT@total@cites}{0}%
  \setcounter{NAT@count@cites}{0}%
  \forcsvlist{\stepcounter{NAT@total@cites}\@gobble}{#3}}{}{}
\newcounter{NAT@total@cites}
\newcounter{NAT@count@cites}
\def\NAT@postnote{}
\def\NAT@hyper@citex#1{%
  \stepcounter{NAT@count@cites}%
  \hyper@natlinkstart{\@citeb\@extra@b@citeb}#1%
  \ifnumequal{\value{NAT@count@cites}}{\value{NAT@total@cites}}
    {\ifNAT@swa\else\if*\NAT@postnote*\else%
     \NAT@cmt\NAT@postnote\global\def\NAT@postnote{}\fi\fi}{}%
  \ifNAT@swa\else\if\relax\NAT@date\relax
  \else\NAT@@close\global\let\NAT@nm\@empty\fi\fi
  \hyper@natlinkend}
\renewcommand\hyper@natlinkbreak[2]{#1}
\patchcmd{\NAT@citex}
  {\ifNAT@swa\else\if*#2*\else\NAT@cmt#2\fi
   \if\relax\NAT@date\relax\else\NAT@@close\fi\fi}{}{}{}
\patchcmd{\NAT@citex}
  {\if\relax\NAT@date\relax\NAT@def@citea\else\NAT@def@citea@close\fi}
  {\if\relax\NAT@date\relax\NAT@def@citea\else\NAT@def@citea@space\fi}{}{}
\makeatother

\usepackage{caption} 

\newcommand{\xhdr}[1]{\noindent{{\bf #1.}}}
\newcommand{\cut}[1]{}

\definecolor{darkpastelgreen}{rgb}{0.01, 0.75, 0.24}

\newcommand{\cmark}{{\color{darkpastelgreen}\ding{51}}}
\newcommand{\xmark}{{\color{red}\ding{55}}}

\title{Do We Still Need Clinical Language Models?}

\author{%
\textbf{Eric Lehman\textsuperscript{1,2}} \quad 
\textbf{Evan Hernandez}\textsuperscript{1, 2} \quad 
\textbf{Diwakar Mahajan}\textsuperscript{3} \quad 
\textbf{Jonas Wulff}\textsuperscript{2} \quad \\
\textbf{Micah J. Smith}\textsuperscript{2} \quad
\textbf{Zachary Ziegler}\textsuperscript{2} \quad 
\textbf{Daniel Nadler}\textsuperscript{2} \quad
\textbf{Peter Szolovits}\textsuperscript{1} \quad \\
\textbf{Alistair Johnson}\textsuperscript{4} \quad
\textbf{Emily Alsentzer}\textsuperscript{5,6}\quad \\
 $^1$MIT \quad 
 $^2$Xyla \quad 
 $^3$IBM Research \quad 
 $^4$The Hospital for Sick Children \quad  \\
 $^5$Brigham and Women's Hospital
 $^6$Harvard Medical School \\
 \texttt{\{lehmer16, dez\}@mit.edu}\\
 }

\begin{document}
\maketitle
\begin{abstract}
Although recent advances in scaling large language models (LLMs) have resulted in improvements on many NLP tasks, it remains unclear whether these models trained primarily with general web text are the right tool in highly specialized, safety critical domains such as \emph{clinical text}.
Recent results have suggested that LLMs encode a surprising amount of medical knowledge. 
This raises an important question regarding the utility of smaller domain-specific language models.
With the success of general-domain LLMs, is there still a need for specialized clinical models? 
To investigate this question, we conduct an extensive empirical analysis of 12 language models, ranging from 220M to 175B parameters, measuring their performance on 3 different clinical tasks that test their ability to parse and reason over electronic health records.
As part of our experiments, we train T5-Base and T5-Large models from scratch on clinical notes from MIMIC III and IV to directly investigate the efficiency of clinical tokens.
We show that relatively small specialized clinical models substantially outperform all in-context learning approaches, even when finetuned on limited annotated data. 
Further, we find that pretraining on clinical tokens allows for smaller, more parameter-efficient models that either match or outperform much larger language models trained on general text. 
We release the code and the models used under the PhysioNet Credentialed Health Data license and data use agreement.\footnote{\url{https://www.physionet.org/content/clinical-t5/1.0.0/}}

\end{abstract}

\section{Introduction}

\begin{figure}
    \centering
    \includegraphics[width=1\textwidth]{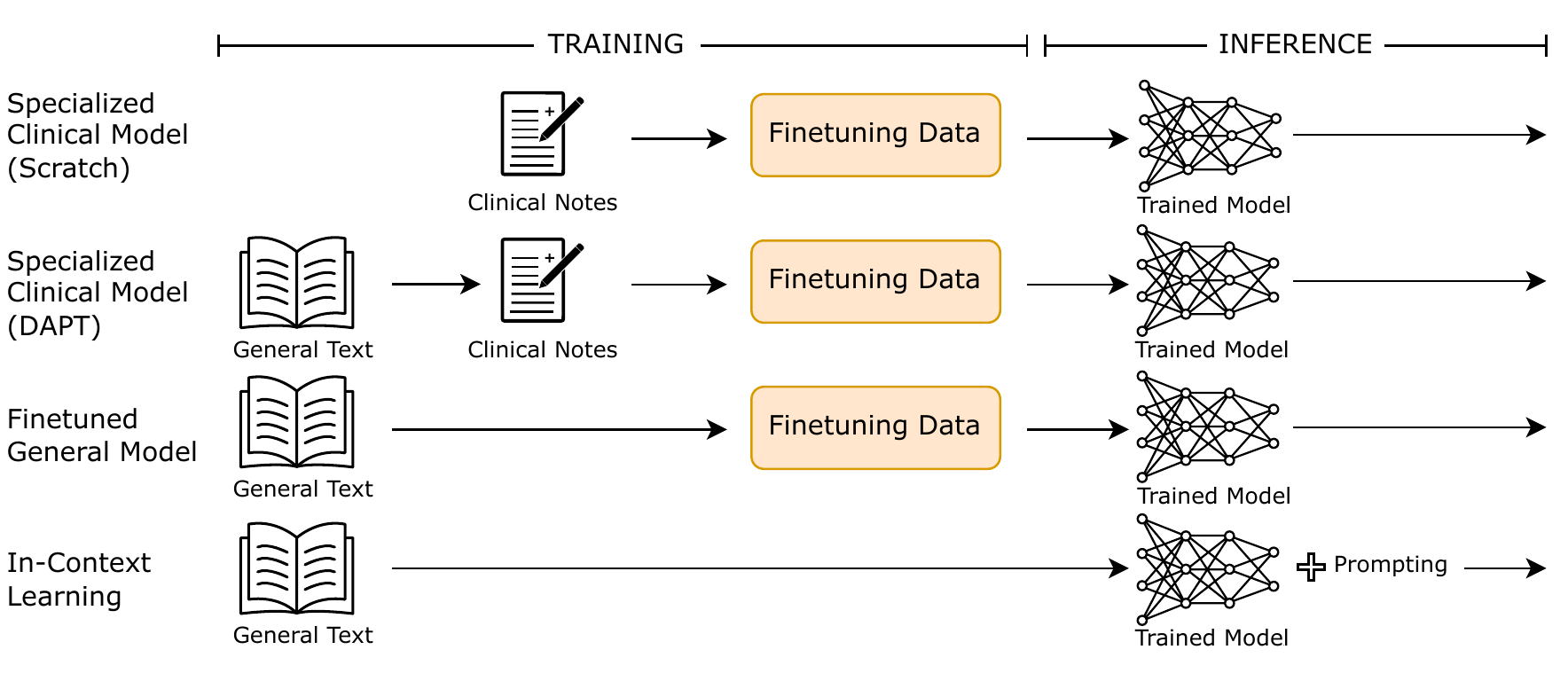}
    \caption{We consider three options for how a healthcare system with access to clinical notes might approach a clinical problem. First, the healthcare system could use a specialized language model pretrained on clinical notes. This model could be pretrained from scratch (Row 1) or from a publicly available checkpoint of a LM pretrained on general text (Row 2).  Alternatively, the healthcare system could directly finetune a publicly available general-purpose language model to perform the clinical task (Row 3). Finally, the healthcare system could use a state-of-the-art LLM such as GPT-3, without any additional finetuning, by prompting the LLM to perform the clinical task (Row 4).}
    \label{fig:options}
\end{figure}





Large language models (LLMs) have shown strong performance on a wide variety of natural language processing (NLP) tasks.
State-of-the-art LLMs are pretrained on billions of tokens scraped from a mixture of general sources, varying widely in both subject matter and quality.
With relatively little task-specific training data, these models can be adapted to new tasks by \textbf{finetuning} the model's weights on labeled data \citep{Devlin2019BERTPO} or by including examples of the task \textbf{in-context} \citep{Kaplan2020ScalingLF, Wei2022EmergentAO}.
This has made them a promising tool for many different applications.

Recent findings have shown that LLMs with over 100B+ parameters contain embedded clinical knowledge \citep{Singhal2022LargeLM}. 
For example, \citet{Agrawal2022LargeLM} found that GPT-3 competes with or outperforms smaller models on a small set of clinical tasks including acronym disambiguation, co-reference resolution, and medication extraction. 
Similarly, ChatGPT achieved passing scores on the US Medical Licensing Exam \citep{chat-gpt-usmle}.
From a performance standpoint, these findings raise an important question about the role of smaller models that are \emph{specifically} tailored for clinical text \citep{alsentzer-etal-2019-publicly, Li2022ClinicalLongformerAC}. 
With the success of LLMs, \textbf{is there still a need for specialized clinical models}? 

To answer this question, we take the perspective of a reasonably equipped healthcare system that is attempting to automate a clinical task involving electronic health record (EHR) notes. For example, suppose a hospital wishes to implement semantic search of clinical notes.
Without automation, a doctor at the hospital would have to manually review all of a patient's previous notes to understand their patient's medical history.
A language model, however, would allow the hospital to automatically extract answers to questions about a patient's medical history, using hundreds of past clinical notes as source material.
A hospital would have three reasonable options for applying a language model to address this type of clinical problem (\Cref{fig:options}):
\begin{enumerate}
    \item Create a \textbf{specialized clinical model} by pretraining a language model on in-house clinical notes and finetuning it for a specific downstream task\footnote{Hospitals could also use a model pretrained on MIMIC.} (\Cref{fig:options}, first and second rows).
    \item Finetune a publicly available pretrained language model, which has largely been pretrained on non-clinical text (\Cref{fig:options}, third row).
    \item Use a state-of-the-art LLM, such as GPT-3, which is made available through an API, and adapt the model to the task using in-context learning (\Cref{fig:options}, last row).
\end{enumerate}

In this paper, we ask whether there is still a need for \textbf{specialized clinical language models}, even with the availability of impressive domain-agnostic LLMs. 
To answer this question, we perform an extensive experimental evaluation of 12 different LMs on 3 different clinical tasks that use EHR notes. 
In addition, we train T5-Base and T5-Large from scratch on clinical notes written primarily in English from the Medical Information Mart for Intensive Care (MIMIC)-III and MIMIC-IV databases \citep{johnson2016mimic, mimic-iv}. 
Our results show that relatively small specialized clinical models (345M parameters) substantially outperform all in-context learning approaches, even when finetuned on limited annotated data. 
We further find that pretraining on clinical tokens allows for smaller, more parameter-efficient models that either match or outperform much larger LMs trained on general text. 
We release the code and models from our experiments under the PhysioNet Credentialed Health Data license and data use agreement.\footnote{Due to the potential for language models to leak protected health information, LLMs trained on clinical datasets such as MIMIC should \emph{not} be released to the general public without evaluating the extent of the leakage. Access to the models requires completion of training in research with human participants (CITI training; \url{https://about.citiprogram.org/series/human-subjects-research-hsr/}) and signing of a data use agreement. Moving forward, we hope to set a precedent for the responsible release of clinical NLP models pretrained or finetuned on MIMIC.}


\section{Background \& Related Work}
\begin{figure}
    \centering
    \includegraphics[width=1.0\textwidth]{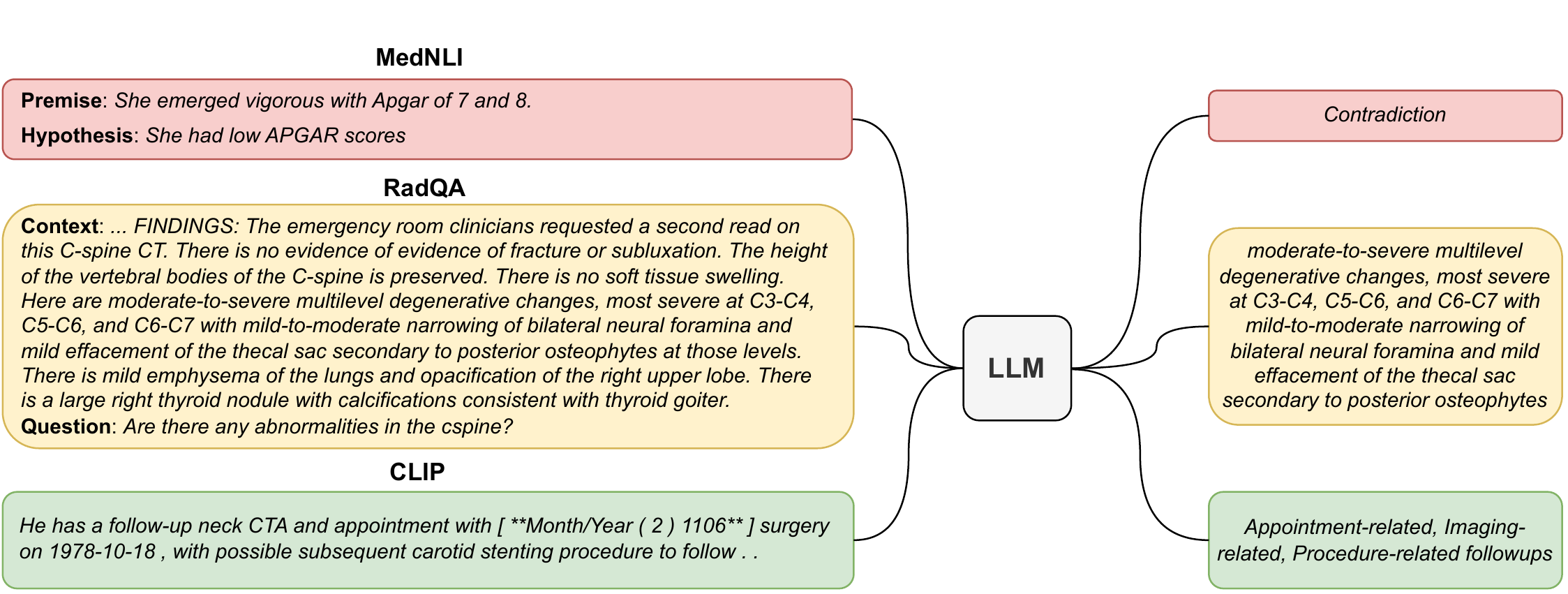}
    \caption{An example of the tasks we consider in this paper. In MedNLI, the goal is determine if the two sentences entail, contradict or are neutral to each other. RadQA is an extractive question answering task over radiology reports. In CLIP, the goal is to identify the different types of patient follow-up information in each sentence of a discharge summary (if any). These examples illustrate the difficulty of parsing clinical text. 
    }
    \label{fig:example}
\end{figure}
We specifically focus on clinical tasks that use EHR notes.
These notes, which are written by clinicians, contain important information about a patient's past medical history, lab results, medications, and current clinical presentation. The text in clinical notes differs substantially from the general-domain text found in LM training corpuses. Some of these differences are highlighted in \Cref{fig:example}: EHR notes often contain grammatical errors (\textit{``no evidence of evidence of fracture"}), include abbreviations not defined in the context (\textit{APGAR}, \textit{CTA}), and reference domain-specific terminology (\textit{carotid stenting}, \textit{subluxation}). These peculiarities also lead to substantial differences between clinical text and biomedical text (such as PubMed). Despite the overall shared domain of medicine, biomedical text is otherwise fluent, edited, and polished.
This makes clinical tasks that involve these notes particularly challenging.
In this section, we briefly describe the three different approaches that one could use for applying a LM to a clinical task (\Cref{fig:options}).

\subsection{Specialized Clinical Models}
We define a \textit{specialized clinical model} to be a model pretrained over clinical notes, and refer to models trained on mostly open-domain web text as \textit{general-purpose models}.
A specialized clinical model can be trained from scratch, or it can be initialized from a previous checkpoint of a biomedical or general-domain model and pretrained further on clinical data in a process known as domain-adaptive pretraining (DAPT, \citealt{Gururangan2020DontSP}). 
Models pretrained on clinical notes have shown improved performance compared to their domain-agnostic equivalents \citep{alsentzer-etal-2019-publicly, lewis-etal-2020-pretrained, Li2022ClinicalLongformerAC, gatortron}.
The semi-structured and abbreviated text found in clinical notes may negatively impact the performance of models pretrained on grammatical biomedical and general text.
Further pretraining on clinical text may help these more general models adapt to this domain-shift.

However, pretraining a LM on clinical notes incurs a high upfront cost. 
This expense may not be justified if it results in only minimal improvements on downstream clinical tasks. 
Additionally, there is a concern that specialized clinical models pretrained on hospital records may retain sensitive patient information \citep{Carlini2018TheSS, Lehman2021DoesBP}. 
For example, \citet{gatortron} train but \emph{do not release} multi-billion parameter models using notes from the University of Florida Health system, likely due to the unknown risk of the models emitting previously seen protected health information.

\subsection{Finetuning General Purpose LLMs for Clinical Tasks} 
As an alternative to pretraining a specialized clinical model, ML practitioners can finetune a general-purpose LM such as the GPT family of models \citep{Radford2018ImprovingLU} or T5 \citep{Raffel2020ExploringTL}, on the clinical task.
The capabilities of these models have been well established in the literature: finetuned general-purpose models are effective at clinical question-answering \citep{Pampari2018emrQAAL}, protected health information (PHI) de-identification \citep{alsentzer-etal-2019-publicly}, and relation-extraction \citep{relation_extraction_bert_success}.
Using a finetuned domain-agnostic model may be necessary in settings where pretraining a language model from scratch is too costly.
While finetuning a general-purpose LM eliminates the cost of pretraining altogether, it may lead to more expensive \emph{inference}-time costs compared to specialized models if the general model must be larger to obtain the same performance. 
Furthermore, these models may still require regular re-finetuning if the data distribution of the EHR shifts, which may happen if, for example, the hospital system changes how medical personnel write notes \citep{Payne2010-si, Blease2020-ne}.
This requires substantially more infrastructure and technical expertise to maintain as model sizes grow. 
There is ongoing research into methods for parameter efficient training \citep{Li2021PrefixTuningOC, Singhal2022LargeLM}, which reduce the computational cost of finetuning. However, in this work, we only consider finetuning of the \emph{entire} model and leave exploration of these techniques to future work.
Finally, in addition to these concerns, models pretrained on text from the general web likely contain additional unexpected and harmful biases towards protected classes and other groups \citep{Bender2021OnTD}.

\subsection{Using In-Context Learning}
A cheaper alternative to finetuning a LM is to use \textbf{in-context learning} (ICL). In this setting, examples of the task are included in the input prompt to the model, and no weights are modified.
ICL has many potential advantages for the clinical domain because there is often a limited set of labeled data due to the high level of expertise needed for annotation.

In-context learning, paired with LLMs like GPT-3, has shown strong performance on a number of tasks \citep{Brown2020LanguageMA}. 
\citet{Agrawal2022LargeLM} found that GPT-3 competes with or outperforms smaller models on several clinical tasks, including acronym disambiguation, co-reference resolution, and medication extraction. 
Due to OpenAI's data policies,\footnote{When \citet{Agrawal2022LargeLM} was released, OpenAI stored all inputs to be used as training data, which violated MIMIC's data use agreement. As of January 16, 2023, it is now possible to use OpenAI models via Microsoft Azure's HIPAA certified platform \citep{azureopenai}.} \citet{Agrawal2022LargeLM} are only able to directly test GPT-3's ability on a restricted set of tasks.
Similarly, \citet{chat-gpt-usmle} found that ChatGPT was able to achieve passing scores on all three stages of the US Medical Licensing Exam (USMLE). 
However, it is unclear whether such performance on tasks requiring clinical knowledge translates to tasks that require parsing semi-structured, abbreviation-laden clinical notes. 

In practice, ICL performs best in very large models \citep{Singhal2022LargeLM} or in models explicitly trained for ICL \citep{Wei2021FinetunedLM}.
These models perform as well as --- or better than --- many finetuned models on several language tasks, which makes ICL a quick and easy option for many NLP problems.
However, GPT-3-scale models are typically accessible only through APIs hosted by private companies, which may add additional concerns about security and data privacy. 
Additionally, these models have a tendency to generate realistic, but factually incorrect content, which may be especially problematic in the safety-critical medical domain.

\section{Experimental Setup}

\begin{table}
\small
\begin{tabular}{lccccc}
\centering
\textbf{Model} & \textbf{Size} & \textbf{Architecture} & \textbf{General PTT} & \textbf{BioMed PTT} & \textbf{Clinical PTT}\\
\toprule 
T5-Base & 220M & Encoder-Decoder & 34B & 0.5B & -- \\
Clinical-T5-Base-Ckpt & 220M & Encoder-Decoder & 34B & 0.5B & 13B \\
Clinical-T5-Base & 220M & Encoder-Decoder & -- & -- & 40B \\
RoBERTa-Large & 345M & Encoder Only & 2200B & -- & -- \\
BioClinRoBERTa & 345M & Encoder Only & -- & 2037B & 65B \\
GatorTron & 345M & Encoder Only & 40B & 92B & 1570B \\
T5-Large & 770M & Encoder-Decoder & 34B & 0.5B & -- \\
Clinical-T5-Large & 770M & Encoder-Decoder & -- & -- & 38B \\
PubMedGPT & 2.7B & Decoder Only & -- & 300B & -- \\
T5-XL & 3B & Encoder-Decoder & 34B & 0.5B & -- \\
Flan-T5-XXL & 11B & Encoder-Decoder & 34B & 0.5B & -- \\
GPT-3 & 175B & Decoder Only & ? & ? & ? \\
\bottomrule
\end{tabular}
\caption{We show all the models used in this paper, as well as their size, architecture and make up of pretraining data. We are unable to provide any information on GPT-3. We focus only on pretraining data, and ignore any finetuning data. PTT stands for pretraining tokens.}

\label{tab:all_models}
\end{table}

We examine the performance of 12 different LMs on three different clinical tasks derived from MIMIC (\Cref{fig:example}). 

\subsection{Tasks}

We select tasks that test the ability to parse and reason over clinical notes. We describe these tasks below:
\begin{itemize}
    \item \textbf{MedNLI} \citep{romanov2018lessons} is a natural language inference task in which the goal is to determine whether a hypothesis written by a doctor can be inferred from a premise taken directly from a clinical note (multi-class classification with labels \textit{entailment}, \textit{neutral}, or \textit{contradiction}).
    We measure performance using accuracy.
    \item \textbf{RadQA} \citep{soni-etal-2022-radqa} is a question-answering (QA) task on radiology reports. Doctors were provided text describing the clinical reason for the imaging and were instructed to ask questions about the radiology report. The answers, if available, were extracted from the report.
    We measure performance using token-level F1 and exact string match metrics.
    \item \textbf{CLIP} \citep{Mullenbach2021CLIPAD} is a multi-label classification task in which the goal is to identify key-sentences that contain some follow-up information in discharge summaries. Each sentence may contain up to 7 possible labels: \texttt{Patient Specific}, \texttt{Appointment}, \texttt{Medication}, \texttt{Lab}, \texttt{Procedure}, \texttt{Imaging}, or \texttt{Other Appointment Related Instructions/Information}.
    We measure performance using micro and macro F1-Score.
\end{itemize}

\subsection{Models}
We experiment with two existing clinical models, BioClinRoBERTa\footnote{We rename the model (RoBERTa-large-PM-M3-Voc) from \citet{lewis-etal-2020-pretrained} to be BioClinRoBERTa.} \citep{lewis-etal-2020-pretrained} and GatorTron \citep{gatortron}, which are both 345M parameter encoder-only models based on the BERT-Large architecture \citep{Devlin2019BERTPO}.
GatorTron was trained on a combination of Wikipedia, PubMed, MIMIC-III, and notes from the University of Florida Health system, whereas BioClinRoBERTa was trained exclusively over PubMed and MIMIC-III. One additional difference between these two models is that GatorTron is trained using the objective function presented in \citet{Lan2019ALBERTAL}, while BioClinRoBERTa is trained using the techniques described in \citet{Liu2019RoBERTaAR}. 

Relative to the general and biomedical domains, there are only a small number of available clinical LMs, primarily due to the paucity of publicly available clinical notes.
To supplement our experiments using \textit{specialized clinical models}, we train three different clinical T5 models on MIMIC III and MIMIC IV, which total $\approx$ 1.2B words (2B tokens).
The T5 models are encoder-decoder LMs that are trained with a generative masked language modeling loss \citep{Devlin2019BERTPO}.
\citet{Raffel2020ExploringTL} pretrain several T5 models of varying size (T5-Base, T5-Large, T5-XL, etc.) on text from the general web. 
We describe our pretrained models below and provide an extensive detail on training method, data preprocessing, and model hyperparameters in \Cref{appendix:preprocessing}:
\begin{itemize}
    \item \textbf{Clinical-T5-Base-Ckpt}: We initialize from the T5-Base (220M) checkpoint and train on MIMIC for 13B tokens. 
    This would classify as a Specialized Clinical Model (DAPT) in row two of \Cref{fig:options}.
    \item \textbf{Clinical-T5-Base}: We initialize T5-Base from scratch and train on MIMIC for 40B tokens. This would classify as a Specialized Clinical Model (Scratch) in row one of \Cref{fig:options}.
    \item \textbf{Clinical-T5-Large}: We initialize T5-Large (770M) from scratch and train on MIMIC for 38B tokens. This would classify as a Specialized Clinical Model (Scratch) in row one of \Cref{fig:options}.
\end{itemize}

To ground the results of the specialized clinical models, we compare to several different general domain models (\Cref{tab:all_models}), including RoBERTa \citep{Liu2019RoBERTaAR}, T5-Base, and T5-Large.
RoBERTa shares the same architecture as GatorTron and BioClinRoBERTa, while T5-Base and T5-Large share the same architecture as Clinical-T5-Base and Clinical-T5-Large, respectively. 
However, RoBERTa, T5-Base and T5-Large are trained exclusively on general-domain text.

In order to examine how specialized clinical models compare to significantly larger, non-clinical models, we compare to PubMedGPT \citep{pubmedgpt} and T5-XL, as these are the largest models that we are able to fully finetune.
All finetuning hyperparameters are reported in \Cref{appendix:model_finetuning}.
Additionally, we examine how these specialized clinical models compare to LLMs used with ICL.
For these experiments, we use GPT-3 (\texttt{text-davinci-003}, \citealt{Ouyang2022TrainingLM}) and T5-Flan-XXL \citep{t5-flan}. 
We explore using a number of different prompts ($\sim$10-20) and report additional details in \Cref{appendix:prompting}.

\section{Clinical Models Are Parameter Efficient}
\label{sec:clinical_efficiency}
\setlength{\tabcolsep}{6pt}
\begin{table}
\centering
{\small
\begin{tabular}{@{}llccccc@{}}
& & MedNLI & \multicolumn{2}{c}{RadQA} & \multicolumn{2}{c}{CLIP} \\
\cmidrule(lr){3-3}
\cmidrule(lr){4-5}
\cmidrule(lr){6-7}
Size & Model &  Acc. & EM & F1 & Micro F1 & Macro F1 \\
\toprule \addlinespace
220M & T5-Base & 0.818 & 0.479 & 0.662 & 0.767 & 0.594 \\
& Clinical-T5-Base-Ckpt & 0.852 & 0.507 & 0.689 & 0.772 & 0.605 \\
& Clinical-T5-Base & 0.855 & 0.531 & 0.710 & 0.793 & 0.652 \\\addlinespace
\toprule \addlinespace
770M & T5-Large & 0.849 & 0.537 & 0.700  & 0.779 & 0.629 \\
& Clinical-T5-Large & \textbf{0.872} & 0.550 & \textbf{0.745} & \textbf{0.800} & \textbf{0.663} \\ \addlinespace
\toprule \addlinespace
3B & T5-XL & 0.869 & \textbf{0.568} & 0.729 & 0.780 & 0.640 \\
\bottomrule
\end{tabular}
\caption{We compare the performance of T5-models with varying pretraining setups. Performance is based on the mean of 3 seeds. Specialized clinical models can outperform larger, general-purpose models like T5-XL.
} 
\label{table:t5_models}
}
\end{table}

In this section, we study how smaller specialized clinical models compare to larger models trained on the general domain.
We fix the \textit{model architecture} and compare models pretrained on general data (T5-Base, T5-Large, T5-XL) versus clinical data (Clinical-T5-Base-Ckpt, Clinical-T5-Base, Clinical-T5 Large).
We find that Clinical-T5-Base-Ckpt and Clinical-T5-Base outperform their general domain counterpart, T5-Base, while Clinical-T5-Large outperforms T5-Large (\Cref{table:t5_models}).
This is despite the fact that we pretrain for several epochs (15+) on the relatively small set of tokens present in MIMIC, which \citet{Raffel2020ExploringTL} shows negatively impacts performance relative to pretraining on unique text for less than one epoch.
Furthermore, we find that pretraining from scratch on clinical data yields the largest performance gains. 
While domain adaptive pretraining of T5-Base on clinical data improves performance over T5-Base, training from scratch is more effective, leading to +3\% and +5\% gains over Clinical-T5-Base-Ckpt on RadQA and CLIP, respectively. 
The weaker performance of Clinical-T5-Base-Ckpt could be explained by a suboptimal learning rate.
Selecting a continuation learning rate is a known challenge of domain-adaptive pretraining \citep{Hoffmann2022TrainingCL}.

While there is substantial evidence that specialized clinical models can outperform their similarly sized general domain equivalents \citep{lewis-etal-2020-pretrained, Liu2019RoBERTaAR, alsentzer-etal-2019-publicly}, it is less clear whether specialized clinical models can outperform \textit{larger} general-domain models. We investigate this by comparing T5 models of varying sizes. We find that Clinical-T5-Base slightly outperforms T5-Large ({3.5$\times$} larger) on all three tasks, but fails to outperform T5-XL ({13.5$\times$} larger). 
Similarly, Clinical-T5-Large slightly outperforms or performs similarly to T5-XL ({3.5$\times$} larger).
This comparison between models trained on in-domain data and larger domain-agnostic models demonstrates that \textbf{specialized clinical models can achieve comparable or better performance with significantly fewer computational resources}. 
This is particularly important for hospital systems, which often lack the infrastructure necessary to run computationally intensive models. 
By training models specifically on in-domain data, hospitals can still benefit from state-of-the-art LLMs, but with a smaller, more manageable model that can operate in computationally constrained environments. 

\subsection{When Is Pretraining From Scratch More Efficient?}
Pretraining a specialized clinical model from scratch has a high initial one-time cost. 
However, performing this pretraining, as our results above suggest, enables the model to be significantly smaller than a general-purpose model while still exhibiting similar downstream performance.
This means that despite a high initial cost, the cost of both finetuning and running inference on a specialized clinical model greatly decreases. In this section, we determine \textbf{at what point} it is more computationally expensive to use a \emph{larger} domain-agnostic model versus pretraining a \emph{smaller} specialized model from scratch.
We measure the cost of a model in terms of FLOPs \citep{Kaplan2020ScalingLF}, which is a function of model size and number of pretraining tokens. We compare the costs of pretraining, finetuning, and performing inference on specialized clinical models versus finetuning and performing inference on an existing general domain model. We assume here that the entire model is updated during the finetuning process.

The training cost $C_{train}$ and inference cost $C_{inf}$ of a model
 are a function of the number of parameters $P$ in the model and the number of tokens $T$ that are processed \citep{Kaplan2020ScalingLF}:
\begin{align}
    C_{train} \left( P, T \right) & = 6 \times P \times T \\
    C_{inf} \left( P, T \right) & = 2 \times P \times T
\end{align}


The number of tokens $T$ in the above cost functions depend on the vocabulary and tokenization process. One additional benefit of training from scratch is that it enables use of an in-domain vocabulary: words previously broken up into word-pieces by a general tokenizer may now be treated as a single token.
We find that for every $1~\texttt{clinical token}$, there are $\approx 1.12~\texttt{general tokens}$.\footnote{We calculate this by running the T5-Base tokenizer over all of MIMIC, as compared to Clinical-T5-Base (same vocabulary size). There is roughly a 65\% overlap between the two vocabularies.}
We model this using an additional token cost weight $w$, with $w_c = 1.0, w_g = 1.12$ for clinical and general-domain tokenizers, respectively. Using $T_{pt}$ pretraining tokens, $T_{ft}$ finetuning tokens (both fixed), and $T_{i}$ inference tokens, we can write the total cost required to pretrain, finetune, and perform inference as follows:

\begin{align} 
C_{model} \left( P, T_i, T_{pt}, T_{ft}, w \right) & = C_{train} \left( P, w T_{pt} \right) + C_{train} \left( P, w T_{ft} \right) + C_{inf} \left( P, w T_{i} \right) \\
& = 6 \times P \times w \times \left( T_{pt} + T_{ft} \right) + 2 \times P \times w \times T_{i}
\end{align}

We can now compare the cost of a small, specialized clinical model of size $P_{clin}$ with a larger, general-domain, previously pretrained (i.e. $T_{pt} = 0$) model of size $P_{gen}$, with $P_{clin} < P_{gen}$. 
Assuming the same amount of finetuning tokens, $T_{ft}$, the costs of both models ($C_{clin}$ and $C_{gen}$) to run inference over $T_i$ tokens becomes:

\begin{align}
C_{clin} \left( P_{clin}, T_{pt}, T_{ft}, T_{i}, w_c \right) & = 6 \times P_{clin} \times w_c \left( T_{pt} + T_{ft} \right) + 2 \times P_{clin} \times w_c T_i \label{eq:cclin} \\
C_{gen} \left( P_{gen}, T_{pt}=0, T_{ft}, T_{i}, w_g \right) & = 6 \times P_{gen} \times w_g T_{ft} + 2 \times P_{gen} \times w_g T_i \label{eq:cgeneral}
\end{align}

Equating \eqref{eq:cclin} and \eqref{eq:cgeneral} and solving for the number of inference tokens, $T_i$, we find the point at which the costs of running inference with the clinical and the general model become equal:

\begin{align}
    T_{i, breakeven} = 
    \frac{3 \left[ w_c P_{clin} \left( T_{pt} + T_{ft} \right) - w_g P_{gen} T_{ft} \right] }
    {w_g P_{gen} - w_c P_{clin} }
    \label{eq:breakeven}
\end{align}

Ignoring finetuning costs and using Clinical-T5-Large and T5-XL as our comparison models, it would take $\sim$40B tokens of inference to recover the costs of pretraining from scratch on clinical data. 
For reference, we estimate that University of Florida Health, which is a large health system with over 1000 beds, records $\sim$15B tokens per year \citep{gatortron}.
While it would take $\sim$2.5 years to recover the cost of a specialized clinical model for a single task that runs over each note once, in practice, such a model would be used for numerous tasks and potentially operate over multiple years of clinical notes.
Given that the two models perform similarly, these results suggest that training a smaller specialized clinical model would allow hospitals to leverage the benefits of LMs, without the higher inference-time and environmental costs of running significantly larger models.

\section{In-Domain Tokens Are More Valuable}
\setlength{\tabcolsep}{5.4pt}
\begin{table}
\small
\begin{center}
\begin{tabular}{@{}llcccccccc@{}}
& & \multicolumn{3}{c}{Compute FLOPs} & \multicolumn{1}{c}{MedNLI}& \multicolumn{2}{c}{RadQA} & \multicolumn{2}{c}{CLIP} \\
\cmidrule(lr){3-5}
\cmidrule(lr){6-6}
\cmidrule(lr){7-8}
\cmidrule(lr){9-10}
Size & Model & General & BioMed & Clinical & Acc. & EM & F1 & Micro & Macro\\
\toprule \addlinespace
220M & T5-Base & 4.5E+19 & 6.6E+17 & -- & 0.818 & 0.479 & 0.662 & 0.767 & 0.594 \\ 
& Clinical-T5-Base & -- & -- & 5.3E+19 & 0.855 & 0.531 & 0.710 & 0.793 & 0.652 \\ 
\addlinespace
\toprule \addlinespace
345M & RoBERTa & 4.6E+21 & -- & -- & 0.852 & 0.521 & 0.684 & 0.793 & 0.677 \\
& BioClinRoBERTa & -- & 4.2E+21 & 1.4E+20 & \textbf{0.900} & \textbf{0.604} & \textbf{0.759} & 0.805 & \textbf{0.707}\\
& GatorTron & 1.4E+19 & 1.9E+20 & 3.3E+21& 0.883 & 0.583 & 0.759 & 0.791 & 0.690\\
\addlinespace
\toprule \addlinespace
770M & T5-Large & 2.6E+19 & 2.3E+18 & -- & 0.849 & 0.537 & 0.700  & 0.779 & 0.629 \\ 
& Clinical-T5-Large & -- & -- & 1.8E+20 & 0.872 & 0.550 & 0.745 & 0.800 & 0.663 \\ \addlinespace
\toprule \addlinespace
2.7B & PubMedGPT & -- & 4.9E+21 & -- & 0.870 & 0.512 & 0.698 & \textbf{0.819} & 0.666\\
3B & T5-XL & 1E+20 & 9E+18 & -- & 0.869 & 0.568 & 0.729 & 0.780 & 0.640\\\addlinespace
\toprule \addlinespace
11B & Flan-T5-XXL & 3.7E+20 & 5.5E+18 & -- & 0.808 & 0.300 & 0.602 & 0.164 & 0.178 \\
175B & GPT-3 & ? & ? & ? & 0.805 & 0.362 & 0.619 & 0.154 & 0.146 \\
\bottomrule
\end{tabular}
\caption{A comparison of clinical and general models trained with varying FLOPs on the three clinical tasks. We only evaluate the ICL methods on 25\% of the test set for CLIP due to the time required for inference on the dataset. We report the mean performance over 3 random seeds. GatorTron and BioClinRoBERTa obtain the highest performance on all metrics except Micro F1 on CLIP. EM stands for exact-match. Macro and Micro stand for Macro and Micro F1 respectively.}
\label{tab:all_results}

\end{center}
\end{table}
In \Cref{sec:clinical_efficiency}, we examine performance based on a \emph{fixed} model architecture. 
In this section, we expand the models we consider to include two more specialized clinical models (GatorTron, BioClinRoBERTa), as well non-clinical models that were trained for a similar number of FLOPs (RoBERTa, PubMedGPT). We aim to explore how performance changes as a function of the amount of general, biomedical and clinical FLOPs used during pretraining. 


BioClinRoBERTa and GatorTron achieve the highest performance on all tasks (\Cref{tab:all_results}).
This is despite the fact that both of these models are less than 12\% of the size of T5-XL, suggesting that model size alone does not guarantee state-of-the-art performance. Another hypothesis is that the total number of FLOPs drives performance; notably, both BioClinRoBERTa and GatorTron were trained for significantly more FLOPs than T5-XL.
However, we find that RoBERTa, which is trained for more total FLOPs than GatorTron and BioClinRoBERTa and shares the same BERT-Large architecture, fails to outperform both of these models. 
This suggests that the high performance of GatorTron and BioClinRoBERTa stems from the makeup of their training data, rather than the total number of FLOPs. 


Similarly, we find that PubMedGPT, which is trained on PubMed for the largest number of total FLOPs, fails to outperform significantly smaller clinical models.
This is especially striking considering that PubMedGPT achieves a high performance on the United States Medical Licensing Exam (USMLE), a set of standardized tests required for medical licensure in the United States  \citep{pubmedgpt}.
In fact, we find that GatorTron scores 10 points \emph{worse} than PubMedGPT on the USMLE, suggesting that there is a difference between the ability to leverage conventional medical knowledge and parse a clinical note.

\begin{figure}
    \centering
    \includegraphics[width=1.0\textwidth]{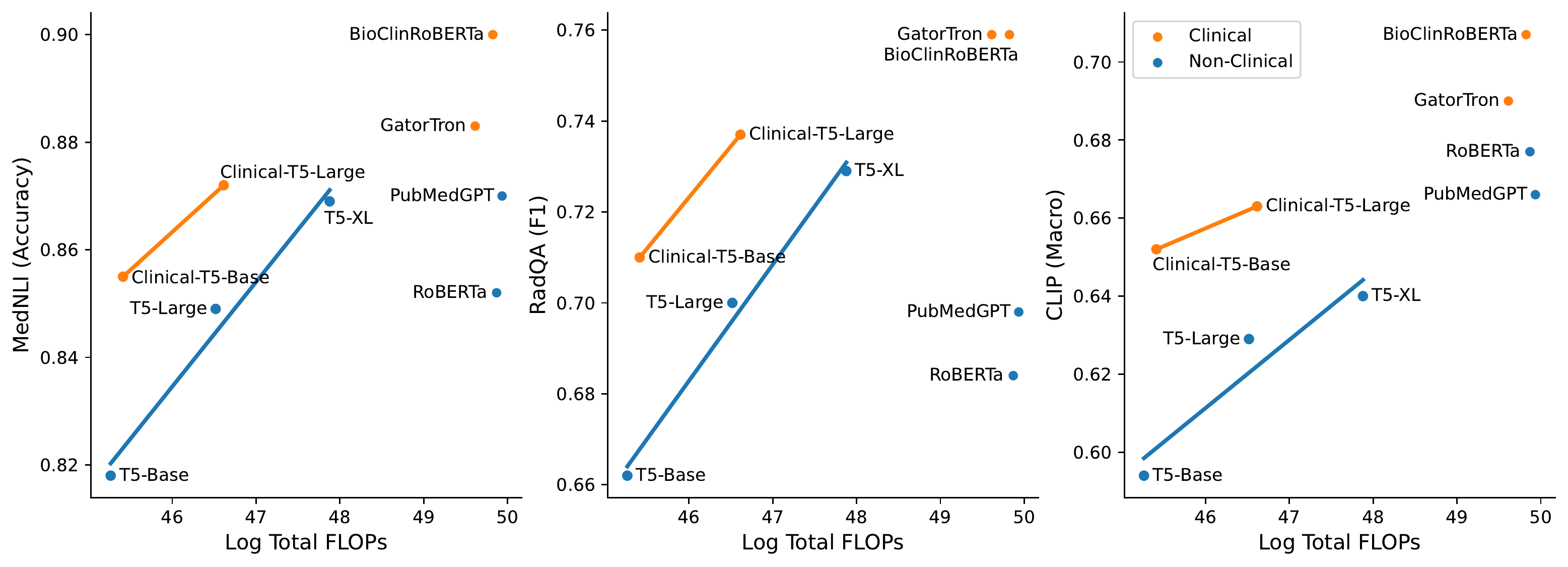}
    \caption{Log total pretraining FLOPs by performance for MedNLI, RadQA, and CLIP. When comparing models with a similar number of FLOPs or performance, clinical models outperform general models. We add regression curves for all T5 models, which are comparable in architecture and training process and differ only in model size and pretraining domain. The T5 models demonstrate the effectiveness of clinical tokens relative to tokens taken from the general web.}
    \label{fig:total_flops}
\end{figure}

As we saw in \Cref{sec:clinical_efficiency}, clinical models outperform their domain-agnostic equivalents. 
\Cref{fig:total_flops} additionally highlights that domain-agnostic models do so with fewer parameters.
Furthermore, given a fixed level of performance, we see that clinical models are more computationally efficient than general-domain models. For example, Clinical-T5-Large and T5-XL achieve comparable performance on MedNLI, yet T5-XL requires 3.5 times as many FLOPs. 
While model architecture differences make a direct comparison difficult, we see that these trends hold for the non-T5 models as well.
These results suggest that \textbf{increasing the number of biomedical and clinical FLOPs, as opposed to the number of parameters or total FLOPs, is the most promising approach for improving performance} on tasks based on clinical text. 

\section{In-Context Learning Underperforms Task Specific Models}
\label{sec:incontext}
\begin{figure}[!htb]
\includegraphics[width=1.00\textwidth]{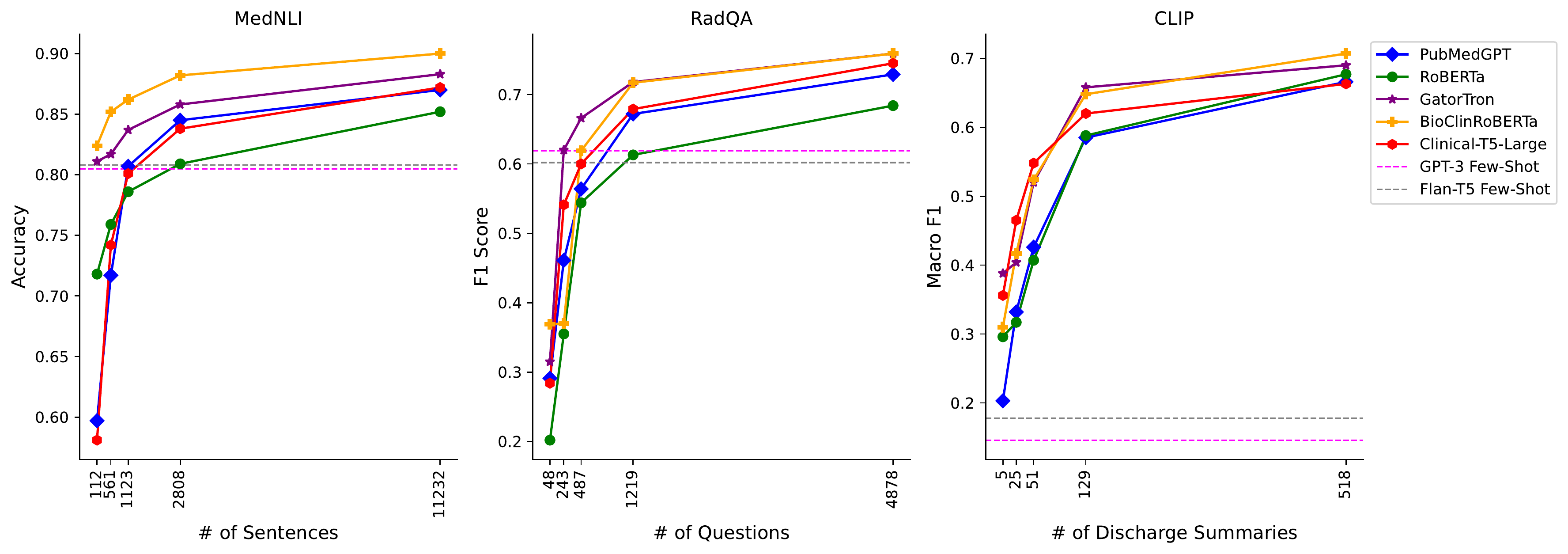}
\caption{An ablation study in which we compare models trained with 1\%, 5\%, 10\%, 25\%, and 100\% of available training data for each task. Except for RadQA at 1\%, GPT-3 and T5-Flan-XXL perform worse than GatorTron at all ablation points. We report mean performance over three random seeds.}
\label{figure:ablation}
\end{figure}
Recent works have shown that LLMs can be adapted to new domains simply through ICL \citep{Wei2022EmergentAO, Lievin2022CanLL, Agrawal2022LargeLM, sanh2021multitask}. 
This type of approach is especially appealing in settings where there is a limited amount of labeled data. 
To properly compare ICL to specialized clinical models and general-purpose models, we simulate a setting in which we have access to very limited data, even as low as $<$ 100 samples. Concretely, we finetune RoBERTa, BioClinRoBERTa, GatorTron, Clinical-T5-Large and PubMedGPT on 1\%, 5\%, 10\%, 25\% and 100\% of the available finetuning data for each task and compare the finetuned models to ICL with GPT-3 and Flan-T5-XXL.

We find that \textbf{models finetuned on all available data significantly outperform any ICL approach} for all of our tasks (\Cref{figure:ablation}). This is consistent with prior results, which compared ICL with parameter-efficient finetuning \citep{Liu2022FewShotPF}.
These findings are particularly relevant to the safety critical clinical domain, where ML practitioners may be willing to gather additional finetuning data for improved performance in high-risk settings.

The utility of specialized clinical models in the few-shot setting varies across datasets. 
On MedNLI, both BioClinRoBERTa and GatorTron outperform GPT-3 in all resource-restricted settings. 
On RadQA, GPT-3 and Flan-T5-XXL outperform the smaller specialized clinical models, but only when the specialized models are trained on 1\% (49 question-answer pairs) of training data. 
It is worth noting that GPT-3 and Flan-T5-XXL are finetuned on question-answering style tasks \citep{Ouyang2022TrainingLM, t5-flan}, albeit it is unlikely that these tasks are from the clinical domain. 

We find that all models outperform GPT-3 and Flan-T5-XXL on CLIP, even when only 5 discharge summaries are used for training data.
We believe that this can be attributed to the aggressive sentence-segmentation of the discharge summaries in the CLIP dataset, as well as the lack of specificity of the task labels.\footnote{The aggressive sentence-segmentation leads to sentences like ``Discharge Instructions:". If important follow-up information follows a header sentence, then the header is also marked with the label of the following sentence. This makes it particularly challenging to do in an ICL setting; however, it is possible that extensive heuristics may help alleviate this issue.}
For example, GPT-3 struggles to categorize labels of type \texttt{Other Appointment Related Instructions}, which significantly lowers its overall performance on CLIP.
Further, unlike RadQA and MedNLI, the label space of this task is different from the type of tasks that GPT-3 and Flan-T5-XXL were finetuned on.

On two of the three datasets, the 11B Flan-T5-XXL model outperforms the much larger 175B GPT-3 model. Flan-T5-XXL is publicly available and can be run with ICL locally on a single GPU, particularly with the aid of libraries such as DeepSpeed \citep{deepspeed}, making it a promising option for ICL when compute is limited. 

We can also examine the gap in performance between clinical (GatorTron, BioClinRoBERTa, Clinical-T5-Large) and non-clinical (RoBERTa, PubMedGPT) pretrained models. 
For RadQA and CLIP in particular, there is a clear gap in performance between clinical and non-clinical models. 
This gap is largest in limited data settings (5\% and 10\%), and slowly diminishes as the amount of finetuning data increases.
This suggests that pretraining on in-domain data can be especially advantageous when there is a low amount of text available for finetuning. 




\section{Limitations \& Future Work}
In this paper, we test 12 different LMs on 3 different clinical tasks. 
We specifically select tasks that test the ability to reason over and parse clinical notes.
However, we do not test the ability of these models to reason over \textit{long text}, which is a considerable challenge when working with clinical notes.
We also do not consider tasks that require generating clinical text (e.g. summarization), which would likely be challenging for encoder-only models.
Further, this work does not consider the various techniques that can be used to reduce model size (e.g., distillation \citep{Hinton2015DistillingTK}, pruning \citep{pruning}) or perform parameter-efficient training (e.g., prompt-tuning \citep{Li2021PrefixTuningOC}). 
Another limitation is that we make some comparisons \textit{across} different architectures.
While this is still a valuable comparison, we cannot attribute improvements in performance to the pretraining data distribution versus the model architecture.
Lastly, we do not use any instruction-tuned models \citep{instruction_tuning}, which are finetuned on a collection of tasks described via instructions, in our finetuning experiments, and we do not compare against ChatGPT, which is not currently available via a HIPAA-certified API.
In the future, we would like to compare to these models and develop instruction-tuned models tailored to the clinical domain.




\section{Conclusion}
In this paper, we explore whether there is still a need for smaller specialized clinical language models. 
To answer this question, we conduct an extensive experimental analysis of 12 models, ranging from 220M to 175B parameters, on 3 different clinical tasks that test the ability to parse and reason over electronic health records.
Our results suggest that smaller models, specifically tailored for clinical text, are more parameter efficient than larger domain-agnostic models.
Further, we find that using in-context learning with extremely large language models, like GPT-3, is not a sufficient replacement for finetuned specialized clinical models.
These findings highlight the importance of developing models for highly specialized domains such as clinical text.

\section{Acknowledgments}
We would like to thank Mark Dredze for his suggestion of using FLOPs to compare between models. We would also like to thank Elena Sergeeva \& Geeticka Chauhan for feedback on an outline of this paper, and Melina Young and Maggie Liu for their help creating figures.
\clearpage
\newpage

\bibliography{main}
\bibliographystyle{apalike}

\newpage
\appendix

\begin{table}
\centering
\begin{tabular}{lcccc}\toprule
\textbf{Name} & \# Patients & \#Notes & \#Words & \\
\hline 
MIMIC-III & 46K & 2M & 429M \\
MIMIC-IV & 246K & 2.6M & 921M \\
MIMIC-III + MIMIC-IV & 291K & 4.1M & 1.2B \\
\bottomrule
\end{tabular}
\caption{We break down the MIMIC-III and MIMIC-IV datasets. There is an overlap in notes between MIMIC-III \& MIMIC-IV.}
\label{table:mimic-breakdown}
\vspace{-2em}
\end{table}

\section{MIMIC Preprocessing and Model Training}
\label{appendix:preprocessing}
In this section, we walk through the steps required to pretrain the T5 specialized clinical models.

\subsection{Data Preprocessing}
We use notes from both MIMIC-III \& MIMIC-IV for pretraining. These datasets are not entirely disjoint, as a portion of the notes that appear in MIMIC-III also appear in MIMIC-IV. However, MIMIC-IV only contains discharge summaries and radiology reports. 
We take the union of MIMIC-III and MIMIC-IV notes such that patient records are not repeated (\Cref{table:mimic-breakdown}). 
This includes notes from all \texttt{CAREVUE} patients and all notes that are not discharge summaries or radiology reports.
We also remove patients that overlap with the tasks we consider in this paper (except for MedNLI). This is important because it is unlikely that models will be pretrained on the same data used at inference time in a realistic deployment scenario.

We remove duplicates of notes from MIMIC-III using \texttt{charttime}, \texttt{storetime} and \texttt{cgid}. 
Duplicate notes can occur when clinicians draft and later edit a note; these duplicates generally differ by 1-2 words.
After this preprocessing, there are 430M words in MIMIC-III (\Cref{table:mimic-breakdown}).

\subsection{Tokenization of DEID Tokens}
All data in MIMIC is fully de-identified. In MIMIC-III, protected health information (PHI) is replaced with special deidentification tags (e.g, \texttt{[**First Name 123**]}), and in MIMIC-IV PHI is replaced with the generic placeholder \texttt{\_\_\_}.
While these de-identification tags can be informative, tokenizers typically break each tag into multiple subwords, dramatically increasing the number of tokens.
We find that replacing all DEID tags with several special DEID tokens (e.g., [NAME]), which we add to the tokenizer vocabulary, reduces the size of MIMIC from 2,400,714,781 tokens to 2,335,573,220 tokens. To perform this replacement on MIMIC-IV, we were granted special access to a file that maps PHI locations to the type of PHI it is. Using this mapping, we add the appropriate DEID tokens to MIMIC-IV text so that the DEID information is stored in a similar manner across both datasets. 

We experimented with 3 different tokenization methods prior to pretraining our specialized clinical models. To select the best tokenizer, we pretrained 3 different models for 10 epochs initializing from T5-Base. 
In the first model, which we use in the paper, we add special DEID tokens and replace the existing ones in MIMIC. 
For the second model, we do not modify the tokenizer at all.
In the last model, we replace all DEID tags with realistic PHI. We frame the problem as a masked language modeling task and query a T5-Large model to generate realistic PHI (e.g. patient names, hospital names, etc.). 
We evaluated each model on the n2c2 2012 challenge \citep{2012_n2c2}, and we found that the performance of these models was comparable. Using the evaluation script provided by \citet{tanl}, we found that n2c2 2012 scores were 0.800, 0.803, 0.802, for the first, second, and third model, respectively.
These models can be made available upon request. 


\subsection{Model Pretraining}
We train and test three different T5 models, following the original T5 training pretraining scheme where possible. We describe the process for training each below.
\begin{enumerate}
    \item Clinical-T5-Base: We pretrain the model from scratch on MIMIC notes for 310K steps, which is roughly 40B tokens worth of pretraining. 
    The model was trained for 200K steps on a TPU before an error with the TPU caused us to switch training to a GPU cluster. The batch size was 32 per TPU/GPU. Due to an issue in the code, the model uses a lowercased vocabulary. All other models are cased.
    \item Clinical-T5-Base-Ckpt: We initialize the model with T5-Base and trained the model for an additional 100K steps on the MIMIC notes.
    The model was trained on 8xA6000 (48GB) GPUs with a batch-size of 32 per GPU. Each epoch took roughly 6 hours. We used 40K warm-up steps (compared to 10K in the original T5 paper) because  we were training the model on a fewer number of tokens. We suspect that this was too many warm-up steps and may have negatively impacted performance.
    \item Clinical-T5-Large: We train this model from scratch on MIMIC notes for 780K steps or approximately 38B tokens. We use a TPU v3.8 cluster with a batch size of 12 per TPU. The cost of training was approximately 1,800 USD, and the training process took approximately 220 hours.
\end{enumerate}

\section{Detailed Model Training and Performance}
\label{appendix:model_finetuning}
\begin{table*}
\small
\begin{tabular}{lcccc|c}\toprule
\textbf{Model} & Size & General PTT & BioMed PTT & Clinical PTT & Unique PTT \\
\hline 
ClinicalBERT & 110M &  137B & 46B & 0.6B & 3.4B / 32B / 0.6B \\
Clinical LongFormer & 150M & 2200B & -- & 15B & 55B / -- / 0.8B \\ 
T5-Base & 220M & 34B & 0.5B & -- & 34B / 0.5B / -- \\
Clinical-T5-Base-Ckpt & 220M & 34B & 0.5B & 13B & 34B / 0.5B / 2.3B \\
Clinical-T5-Base & 220M & -- & -- & 40B & -- / -- / 2B \\
RoBERTa-Large & 345M & 2200B & -- & -- & 55B / -- / -- \\
BioClinRoBERTa & 345M & -- & 2037B & 65B & -- / 32B / 0.8B \\
GatorTron & 345M & 40B & 92B & 1570B & 4B / 9B / 157B \\
T5-Large & 770M & 34B & 0.5B & -- & 34B / 0.5B / -- \\
Clinical-T5-Large & 770M& -- & -- & 38B & -- / -- / 2B \\
SciFive & 220M & 34B & 27B & -- &  34B / 27B / -- \\
SciFive-Large & 770M & 34B & 14B & -- & 34B / 14B / --\\
PubMedGPT & 2.7B & -- & 300B & -- & -- / 50B / -- \\
T5-XL & 3B & 34B & 0.5B & -- & 34B / 0.5B / -- \\
Flan-T5-XXL & 11B & 34B & 0.5B & -- & 34B / 0.5B / -- \\
GPT-3 & 175B & -- & -- & -- & -- \\
\bottomrule
\end{tabular}
\caption{PTT stands for pretraining tokens. All of the models tested and considered for the project. We show the models, their size, what they were initialized from, and the make up of their pretraining data. We are, of course, unable to provide any information on GPT-3. We focus only on pretraining data, and ignore any instruction tuning data.}
\end{table*}

In the following section, we describe our process for finetuning language models on MedNLI, RadQA, and CLIP. Due to space limitations, we only show results for 12 models in the main body of the paper. However, in this expanded appendix, we report the performance of 16 different general, biomedical, and clinical language models, adding results for ClinicalBERT \citep{alsentzer-etal-2019-publicly}, ClinicalLongformer \citep{Li2022ClinicalLongformerAC}, SciFive \citep{Phan2021SciFiveAT}, and SciFive Large.
All of these models were trained use DAPT. 
ClinicalBERT was initialized from BioBERT and further pretrained over MIMIC-III.
Similarly, ClinicalLongformer was initialized from 
 the Longformer \citep{Beltagy2020LongformerTL} and trained over MIMIC-III.
 Lastly, SciFive and SciFive-Large were initialized from T5-Base and T5-Large, respectively, and trained over PubMed.

\subsection{Hyperparameter Tuning}
We largely follow the guidance of \citet{Raffel2020ExploringTL} for finetuning all of the T5 models. \citet{Raffel2020ExploringTL} suggest using a constant learning rate of 1e-3 for all finetuning experiments (with adafactor optimizer).We found that this was too large and that 1e-4 performed significantly better across all tasks. 

For PubMedGPT, we follow \citet{pubmedgpt} and train using AdamW with a learning rate of 2e-6. We experimented with 2e-5, but found that 2e-6 performed much better. For ClinicalBERT, GatorTron, and ClinicalLongformer, we do a hyperparameter search over learning rates of 2e-5, 3e-5 and 5e-5.
For RoBERTa and BioClinRoBERTa, we follow the guidence of \citet{lewis-etal-2020-pretrained}, and use a learning rate of 1e-5. We select whichever learning rate works best on the validation set. The optimal learning rate varies for each task. We use the AdamW optimizer \citep{Loshchilov2017FixingWD}.

To train T5-XL and PubMedGPT with limited GPU resources, we leverage the DeepSpeed library \citep{deepspeed}. This enables the models to be trained on 32GB GPUs by using CPU offloading at the expense of increasing train run time. 

We train until convergence for all tasks. The time to convergence differs across tasks. Generally, we find that T5-XL converges much faster than the other T5 models. 
On MedNLI, for example, T5-XL converges within 15 epochs whereas Clinical-T5-Large needs roughly 30-40 epochs to converge. We ran all experiments with an effective batch size of 64.
We select the optimal hyperparameters according to the performance on the vaidation set for each task (accuracy for MedNLI, F1 for RadQA, and Macro F1 for CLIP).

\subsection{Computational Resources and Run-Time}
We used a wide-range of GPUs for our experiments, including 80GB V100s, 48GB A6000, 32GB V100, and 12GB 2080Tis. The encoder-only models take around 20-40 minutes to run on MedNLI and RadQA and 3 hours to run on CLIP. We find that the T5-Base models take around an hour to run on MedNLI and RadQA and 4 hours on CLIP (these models are trained for additional epochs compared to the encoder-only models because they are slower to converge). The T5-Large models take around 1.5 hours to run on MedNLI and RadQA and roughly 10 hours to run on CLIP. PubMedGPT and T5-XL take around 6 hours to run on MedNLI and RadQA. For CLIP, this took roughly 40 hours to run (on 4x48GB GPUs). The use of the DeepSpeed library increased the time required for finetuning PubMedGPT and T5-XL.

\subsection{Task-Specific Details}
We produce answers with the T5 models by generating the label or extracted text with beam search. For the encoder-only models and PubmedGPT, we add a task-specific linear layer on top of the base model.
We next outline finetuning details that are specific to each task. 

\xhdr{MedNLI}
We train the encoder-only models and PubMedGPT for 20 epochs, and we train T5-XL for 15 epochs. 
All clinical and general-domain T5-Base and T5-Large models are trained for 40 epochs. For all T5 models, we use a beam search width of 3.

\xhdr{RadQA}
As before, we train the encoder-only models and PubMedGPT for 20 epochs, and we train T5-XL for 15 epochs. We trained all T5-Base and T5-Large models for 50 epochs. 
For all T5 models, we use a beam search width of 1. We found that increasing the beam-search width did not consistently improve performance; we experimented with beam search widths of 3, 5, and 10, and found that it increased exact-match at the expense of F1-Score.

\xhdr{CLIP}
Again, we train the encoder-only models and PubMedGPT for 20 epochs, and we train T5-XL for 15 epochs. We trained all T5-Base and T5-Large models for 40 epochs. For all T5 models, we use a beam search width of 5. 
We did not experiment with different beam search widths for CLIP.
To generate multiple labels for each sentence, we ask the T5 models to produce a comma-delimited list of labels, ordered alphabetically. 
We use a context window of 256 for all experiments with CLIP. This resulted in a slightly lower performance compared to the results presented in \citet{Mullenbach2021CLIPAD}, which used a window of 512 tokens.

\begin{table*}
\small
\begin{tabular}{lccccccc}\toprule
\textbf{Task} & Type & Labels & Max Sequence Length & Train / Val / Test & Units \\
\hline 
MedNLI & NLI & 3 & 256 & 11K / 1K / 1K & Sentence Pairs\\
RadQA & QA & -- & 1024 & 4.8K / 1K / 1K & Question + Answer Pairs\\
CLIP & CLS  & 7 & 256 & 107K / 10K / 10K & Sentences\\
\bottomrule
\end{tabular}
\caption{We summarize some task statistics. CLS stands for classification.}
\end{table*}

\section{Additional Discussion of Model Performance}

\begin{table*}
\centering
\begin{tabular}{lccccc}\toprule
\textbf{Model} & \textbf{Size} & \textbf{BioMed PT} & \textbf{Clinical PT} & \textbf{Accuracy} & \textbf{Std.} \\
ClinicalBERT & 110M & \xmark & \cmark & 0.815 & 0.008\\
ClinicalLongFormer & 150M & \xmark & \cmark & 0.846 & 0.003 \\
T5-Base & 220M & \xmark & \xmark & 0.818 & 0.006 \\
SciFive & 220M & \xmark & \xmark & 0.835 & 0.003 \\
Clinical-T5-Base-Ckpt & 220M & \xmark & \cmark & 0.852 & 0.007 \\
Clinical-T5-Base & 220M & \xmark & \cmark & 0.855 & 0.004 \\
GatorTron & 345M & \cmark & \cmark & 0.883 & 0.002 \\
RoBERTa  & 345M & \xmark & \xmark & 0.852 & 0.002 \\
BioClinical RoBERTa & 345M & \cmark & \cmark & 0.900 & 0.003 \\T5-Large & 770M & \xmark & \xmark & 0.849 & 0.008 \\
SciFive Large & 770M & \cmark & \xmark & 0.857 & 0.005 \\
Clinical-T5-Large & 770M & \xmark & \cmark & 0.872 & 0.008  \\
PubmedGPT & 2.7B & \cmark & \xmark &  0.870 & 0.009  \\
T5-XL & 3B & \xmark & \xmark & 0.869 & 0.004 \\
Flan-T5-XL & 11B & \xmark & \xmark & 0.808 & --\\ 
GPT-3 & 175B & -- & -- & 0.807 & -- \\
\bottomrule
\end{tabular}
\caption{We show the performance of all models considered on MedNLI. Results are based on at least 3 seeds.}
\label{appendix:all_mednli}
\end{table*}
\subsection{MedNLI} We report results for all models in \Cref{appendix:all_mednli}.
We find that ClinicalBERT performs similarly to T5-Base, while  ClinicalLongFormer performs similarly to T5-Large.
We additionally test SciFive and SciFive-Large \citep{Phan2021SciFiveAT}, which outperform T5-Base and T5-Large, respectively.
However, these models fail to outperform Clinical-T5-Base and Clinical-T5-Large.
This may be because SciFive and SciFive-Large are trained via DAPT, while Clinical-T5-Base and Clinical-T5-Large are trained from scratch.
Further, SciFive and SciFive-Large are trained on biomedical tokens, rather than clinical tokens. 

We also show how performance changes depending on the number of DAPT steps (\Cref{appendix:tab_chkpting_scores}). 
We find that training Clinical-T5-Base-Ckpt for 20K pretraining steps gives a reasonable boost in performance over T5-Base.
Training from 20K to 80K steps does not seem to provide any additional performance gains. 
However, we find that training for 100K steps does improve performance versus training for 80K steps.
This is likely due to the learning rate scheduler. 
It is possible that at 40K to 80K steps, the learning rate is too large.

\begin{table}
\centering
\begin{tabular}{lccc}\toprule
\textbf{Model} & \textbf{Clinical PTT} & \textbf{Accuracy} & \textbf{Std.} \\
T5-Base & -- & 0.818 & 0.006 \\
Clinical-T5-Base-Ckpt-20K & 2B & 0.831 & 0.001 \\
Clinical-T5-Base-Ckpt-40K & 5B & 0.831 & 0.002 \\
Clinical-T5-Base-Ckpt-60K & 8B & 0.836 & 0.007 \\
Clinical-T5-Base-Ckpt-80K & 10B & 0.836 & 0.002\\
Clinical-T5-Base-Ckpt & 13B & 0.852 & 0.007\\
\bottomrule
\end{tabular}
\caption{We report the performance of Clinical-T5-Base-Ckpt on MedNLI when trained on an increasing number of tokens from MIMIC. We find that pretraining for a high warmup initially boosts performance by 1\%.}
\label{appendix:tab_chkpting_scores}
\end{table}

\begin{table}
\small
\centering
\begin{tabular}{lccccc} \toprule
\textbf{Model} & \textbf{Size} & \textbf{BioMed PT} & \textbf{Clinical PT} & \textbf{Exact Match} & \textbf{F1} \\ 
ClinicalBERT & 110M & \xmark & \cmark & 0.457 $\pm$ 0.002 & 0.626 $\pm$ 0.008 \\
ClinicalLongformer & 150M & \xmark & \cmark & 0.518 $\pm$ 0.036 & 0.689 $\pm$ 0.018\\
T5-Base & 220M & \xmark & \xmark & 0.479 $\pm$ 0.014 & 0.662 $\pm$ 0.010 \\
SciFive & 220M & \cmark & \cmark & 0.506 $\pm$ 0.010 & 0.697 $\pm$ 0.007 \\
Clinical-T5-Base-Ckpt & 220M & \xmark & \cmark & 0.505 $\pm$ 0.014 & 0.684 $\pm$ 0.009 \\
Clinical-T5-Base & 220M & \xmark & \cmark & 0.531 $\pm$ 0.013 & 0.710 $\pm$ 0.005 \\
RoBERTa & 345M & \xmark & \xmark & 0.521 $\pm$ 0.014 & 0.684 $\pm$ 0.004 \\ 
BioClinical RoBERTa & 345M & \xmark & \xmark & 0.604 $\pm$ 0.012  & 0.759 $\pm$ 0.029 \\ 
GatorTron & 345M & \cmark & \cmark & 0.583 $\pm$ 0.008 & 0.759 $\pm$ 0.008 \\
T5-Large & 770M & \xmark & \xmark & 0.537 $\pm$ 0.019 & 0.700 $\pm$ 0.012 \\
SciFive-Large & 770M & \cmark & \xmark & 0.541 $\pm$ 0.016 & 0.704 $\pm$ 0.013 \\
Clinical-T5-Large & 770M & \xmark & \cmark & 0.550 $\pm$ 0.018 & 0.745 $\pm$ 0.008 \\
PubMedGPT & 2.7B & \cmark & \xmark & 0.512 $\pm$ 0.005 & 0.698 $\pm$ 0.004 \\
T5-XL & 3B & \xmark & \xmark & 0.568 $\pm$ 0.007 & 0.729 $\pm$ 0.005 \\
Flan-T5-XXL & 11B & \xmark & \xmark & 0.300 & 0.602 \\ 
GPT-3 & 175B & \xmark & \xmark & 0.362 & 0.620 \\ 
\bottomrule
\end{tabular}
\caption{Performance of all models on RadQA.  We report the mean performance and standard deviation of models trained with at least 3 random seeds.}
\label{appendix:all_radqa}
\end{table}
\subsection{RadQA} We report results for all models in \Cref{appendix:all_radqa}. We find that ClinicalBERT performs extremely poorly on RadQA, while the ClinicalLongformer performs similar to Clinical-T5-Base-Ckpt. Similar to MedNLI, SciFive and SciFive-Large outperform T5-Base and T5-Large, respectively.
However, both of these models fail to outperform their clinical equivalents. 

\begin{table}
\centering
\begin{tabular}{lccccc} \toprule
\textbf{Model} & \textbf{Size} & \textbf{BioMed PT} & \textbf{Clinical PT} & \textbf{Micro F1} & \textbf{Macro F1} \\ 
ClinicalBERT & 110M & \xmark & \cmark & 0.777 $\pm$ 0.006 & 0.649 $\pm$ 0.007 \\
ClinicalLongformer & 150M & \xmark & \cmark & 0.790 $\pm$ 0.003 & 0.659 $\pm$ 0.008\\
T5-Base & 220M & \xmark & \xmark & 0.767 $\pm$ 0.008 & 0.594 $\pm$ 0.011 \\
SciFive & 220M & \cmark & \cmark & 0.769 $\pm$ 0.008 & 0.603 $\pm$ 0.004 \\
Clinical-T5-Base-Ckpt & 220M & \xmark & \cmark & 0.772 $\pm$ 0.005 & 0.605 $\pm$ 0.009 \\
Clinical-T5-Base & 220M & \xmark & \cmark & 0.793 $\pm$ 0.001 & 0.652 $\pm$ 0.009 \\
RoBERTa & 345M & \cmark & \xmark & 0.793 $\pm$ 0.001 & 0.677 $\pm$ 0.008 \\
BioClinRoBERTa & 345M & \cmark & \xmark &  0.805 $\pm$ 0.005 & 0.707 $\pm$ 0.007 \\
GatorTron & 345M & \cmark & \xmark & 0.791 $\pm$ 0.003 & 0.690 $\pm$ 0.010\\
T5-Large & 770M & \xmark & \xmark & 0.779 $\pm$ 0.008 & 0.629 $\pm$ 0.011 \\
SciFive-Large & 770M & \cmark & \xmark & 0.774 $\pm$ 0.008 & 0.630 $\pm$ 0.011 \\
Clinical-T5-Large & 770M & \xmark & \cmark & 0.800 $\pm$ 0.008 & 0.663 $\pm$ 0.007 \\
PubMedGPT & 2.7B & \cmark & \xmark & 0.819 $\pm$ 0.003 & 0.666 $\pm$ 0.003 \\
T5-XL & 3B & \xmark & \xmark & 0.780 $\pm$ 0.021 & 0.640 $\pm$ 0.022 \\
Flan-T5-XXL & 11B & \xmark & \xmark & 0.164 & 0.178 \\
GPT-3 & 175B & \xmark & \xmark & 0.154 & 0.146 \\ 
\bottomrule
\end{tabular}
\caption{Performance of all models on CLIP. We report the mean performance and standard deviation of models trained with at least 3 random seeds. T5-Flan-XXL and GPT-3 are based on a sample of 25\% of the test data.}
\label{appendix:all_clip}
\end{table}

\subsection{CLIP} We report results for all models in \Cref{appendix:all_clip}. 
We find that ClinicalBERT and ClinicalLongformer perform very well on this task, performing comparably to or outperforming the much larger T5-XL model. This is likely due to the fact that the the T5 models \textit{generate} answers, which is challenging for a multi-label classification task. As we saw in other experiments, SciFive and SciFive-Large underperform their clinical-domain counterparts. PubMedGPT has the highest Micro F1 performance, outperforming both GatorTron and BioClinRoBERTa, which excelled across all other tasks. 

\section{Additional Details about In Context Learning Experiments}
\label{appendix:prompting}
In this section, we provide additional information about our approach for performing in context learning with GPT-3 and Flan-T5-XXL. 

We experiment with approximately 5-10 different prompts for each task, crafting prompts to reflect the prompts used during instruction tuning of Flan-T5 and GPT-3. We pair each prompt with one to three randomly sampled examples for in-context learning. We select the best prompt based on the performance on a random sample of 200 examples from the validation set.
We use a temperature of 0 and a beam search width of 1.


There are two options for generating labels for CLIP, which is a multi-label classification task. The model can either generate predictions for each label independently or all at once. We experiment with both options using Flan-T5-XXL and find that both approaches perform similarly. However, 
independently prompting the model for each label results in higher inference time costs. Therefore, we ask the model to generate predictions for all labels at once for GPT-3. 

We list the prompts that were used on the test set below. Note that we only include the prompt itself and do not include the in-context examples.
\begin{itemize}
    \item MedNLI - T5-Flan-XXL \& GPT-3: \texttt{Answer entailment, contradiction or neutral. Premise: \{Premise\} Hypothesis: \{Hypothesis\}}
    \item RadQA - GPT-3 \& GPT-3: \texttt{Context: \{Context\}, \{Question\} Answer N/A if there is no answer or give a quote from the context:}
    \item CLIP - T5-Flan-XXL: 
    \begin{enumerate}
        \item \texttt{Context: \{Context\}. Does the above sentence contain information about current or future appointments? Options: -Yes -No}
        \item \texttt{Context: \{Context\}. Does the above sentence contain information about medications? Options: -Yes -No }
        \item \texttt{Context: \{Context\}. Does the above sentence contain any important actionable information?  Options: -Yes -No }
        \item \texttt{Context: \{Context\}. Does the above sentence contain any information about laboratory tests? Options: -Yes -No }
        \item \texttt{Context: \{Context\}. Does the above sentence contain any information about what to do post-discharge?  Options: -Yes -No }
        \item \texttt{Context: \{Context\}. Does the above sentence contain any information about procedures (e.g., surgeries)?  Options: -Yes -No }
        \item \texttt{Context: \{Context\}. Does the above sentence contain any information about an imaging followup? Options: -Yes -No }
    \end{enumerate}
    \item CLIP - GPT-3: \texttt{Context: \{Context\}. Label the above sentence as one or more of the following, delimited by comma: Options: -Appointment-related followup information -Medication-related followup information -Lab-related followup information -Case-specific instructions for the patient -Procedure-related followup information -Imaging-related followup information -None of the above }
\end{itemize}

We will make all of our prompts available, along with their validation set performance scores. 
Consistent with prior literature, we find that the performance of these models is extremely dependent on the prompt \citep{t5-flan}. 
For example, when evaluating Flan-T5-XXL on MedNLI, we find that using the following prompt leads to a drop in accuracy from 83.5\% to 62\% on the validation set: \texttt{Answer entailment, neutral or contradiction. Premise: {Premise} Hypothesis: {Hypothesis}. Answer:'}.

Post-processing was required to map the text generated by GPT-3 and Flan-T5-XXL to the label space. For MedNLI, we check if the string contains the word entailment, contradiction or neutral. If none of these three words appear, we predict neutral. For CLIP, we search the generated string for the label types. This allows for the models to generate predictions in any order. GPT-3 and Flan-T5-XXL sometimes produce answers to RadQA questions that cannot be extracted directly from the radiology report. In such cases, we calculate F1-score regardless.
Had we enforced that the model produce a string directly from the text, the F1-score would have dropped to $\sim$40 for both models.

Finally, we report the exact performance metrics shown in \Cref{figure:ablation} in \Cref{appendix:ablation-mednli}, \Cref{appendix:ablation-radqa-f1} and \Cref{appendix:ablation-clip-macro}. 
We also report Exact Match on RadQA in \Cref{appendix:ablation-radqa-em} and Micro F1 on CLIP in \Cref{appendix:ablation-clip-micro}.
We initially experimented with GPT-Neo-X \citep{gpt-neox-20b} in addition to GPT-3 and T5-Flan-XXL. However, in our initial experiments, we found that its performance on MedNLI was less than 40\%. Therefore, we dropped it from our remaining experiments.

\begin{table}[!ht]
    \centering
    \small
    \begin{tabular}{|l|l|l|l|l|l|}
    \hline
        Model & 1\% & 5\% & 10\% & 25\% & 100\% \\ \hline
        PubMedGPT & 0.597 +/- 0.011 & 0.717 +/- 0.011 & 0.807 +/- 0.011 & 0.845 +/- 0.006 & 0.870 +/- 0.009 \\ \hline
        GatorTron & 0.811 +/- 0.001 & 0.817 +/- 0.005 & 0.837 +/- 0.023 & 0.858 +/- 0.001 & 0.883 +/- 0.002 \\ \hline
        RoBERTa & 0.718 +/- 0.008 & 0.759 +/- 0.010 & 0.786 +/- 0.008 & 0.809 +/- 0.004 & 0.852 +/- 0.002 \\ \hline
        BioClinRoBERTa & 0.824 +/- 0.025 & 0.852 +/- 0.004 & 0.862 +/- 0.004 & 0.882 +/- 0.006 & 0.900 +/- 0.003 \\ \hline
        Clinical-T5-Large & 0.581 +/- 0.029 & 0.742 +/- 0.033 & 0.801 +/- 0.003 & 0.838 +/- 0.007 & 0.872 +/- 0.008 \\ \hline
    \end{tabular}
    \caption{Accuracy on MedNLI for models finetuned with varying amounts of annotated data. Percentages refer to fraction of the training set for the task. We report the mean and standard deviation over three random seeds. We always evaluate on the full test set.}
    \label{appendix:ablation-mednli}
\end{table}

\begin{table}[!ht]
    \centering
    \small
    \begin{tabular}{|l|l|l|l|l|l|l|l|l|l|}
    \hline
        Model & 1\% (F1) & 5\% (F1) & 10\% (F1) & 25\% (F1) & 100\% (F1) \\ \hline
        PubMedGPT & 0.291 +/- 0.017 & 0.461 +/- 0.002 & 0.564 +/- 0.012 & 0.672 +/- 0.014 & 0.729 +/- 0.005 \\ \hline
        GatorTron & 0.315 +/- 0.027 & 0.620 +/- 0.011 & 0.666 +/- 0.001 & 0.718 +/- 0.008 & 0.759 +/- 0.008  \\ \hline
        RoBERTa & 0.202 +/- 0.014 & 0.355 +/- 0.015 & 0.544 +/- 0.006 & 0.613 +/- 0.008 & 0.684 +/- 0.004 \\ \hline
        BioClinRoBERTa & 0.369 +/- 0.001 & 0.370 +/- 0.011 & 0.619 +/- 0.021 & 0.717 +/- 0.011 & 0.759 +/- 0.029 \\ \hline
        Clinical-T5-Large & 0.284 +/- 0.024 & 0.541 +/- 0.027 & 0.600 +/- 0.021 & 0.679 +/- 0.012 & 0.745 +/- 0.008 \\ \hline
    \end{tabular}
    \caption{F1 score on RadQA for models finetuned with varying amounts of annotated data. Percentages refer to fraction of the training set for the task. We report the mean and standard deviation over three random seeds. We always evaluate on the full test set.}
    \label{appendix:ablation-radqa-f1}

\end{table}

\begin{table}[!ht]
    \centering
    \small
    \begin{tabular}{|l|l|l|l|l|l|}
    \hline
        Model & 1\% (EM) & 5\% (EM) & 10\% (EM) & 25\% (EM) & 100\% (EM) \\ \hline
        PubMedGPT & 0.231 +/- 0.004 & 0.332 +/- 0.012 & 0.362 +/- 0.009 & 0.476 +/- 0.013 & 0.512 +/- 0.005 \\ \hline
        GatorTron & 0.263 +/- 0.022 & 0.482 +/- 0.010 & 0.507 +/- 0.004 & 0.554 +/- 0.012 & 0.583 +/- 0.008  \\ \hline
        RoBERTa & 0.187 +/- 0.021 & 0.295 +/- 0.004 & 0.415 +/- 0.009 & 0.462 +/- 0.009 & 0.521 +/- 0.014 \\ \hline
        BioClinRoBERTa & 0.322 +/- 0.009 & 0.322 +/- 0.009 & 0.479 +/- 0.016 & 0.561 +/- 0.019 & 0.604 +/- 0.012 \\ \hline
        Clinical-T5-Large & 0.206 +/- 0.015 & 0.358 +/- 0.016 & 0.435 +/- 0.024 & 0.495 +/- 0.006 & 0.550 +/- 0.018 \\ \hline
    \end{tabular}
    \caption{Exact Match performance on RadQA for models finetuned with varying amounts of annotated data. Percentages refer to fraction of the training set for the task. We report the mean and standard deviation over three random seeds. We always evaluate on the full test set.}
    \label{appendix:ablation-radqa-em}
\end{table}

\begin{table}[!ht]
    \small
    \centering
    \begin{tabular}{|l|l|l|l|l|l|}
    \hline
        Model & 1\% (Micro) & 5\% (Micro) & 10\% (Micro) & 25\% (Micro) & 100\% (Micro) \\ \hline
        PubMedGPT & 0.580 +/- 0.006 & 0.706 +/- 0.010 & 0.740 +/- 0.006 & 0.789 +/- 0.003 & 0.819 +/- 0.003 \\ \hline
        GatorTron & 0.686 +/- 0.010 & 0.725 +/- 0.009 & 0.759 +/- 0.006 & 0.785 +/- 0.002 & 0.793 +/- 0.001 \\ \hline
        RoBERTa & 0.703 +/- 0.014 & 0.726 +/- 0.002 & 0.739 +/- 0.001 & 0.768 +/- 0.006 & 0.791 +/- 0.003 \\ \hline
        BioClinRoBERTa & 0.692 +/- 0.007 & 0.714 +/- 0.003 & 0.739 +/- 0.003 & 0.770 +/- 0.001 & 0.805 +/- 0.005\\ \hline
        Clinical-T5-Large & 0.616 +/- 0.004 & 0.716 +/- 0.016 & 0.743 +/- 0.013 & 0.777 +/- 0.000 & 0.800 +/- 0.008 \\ \hline
    \end{tabular}
    \caption{Micro F1 score on CLIP for models finetuned with varying amounts of annotated data. Percentages refer to fraction of the training set for the task. We report the mean and standard deviation over three random seeds. We always evaluate on the full test set.}
    \label{appendix:ablation-clip-micro}

\end{table}

\begin{table}[!ht]
    \centering
    \small
    \begin{tabular}{|l|l|l|l|l|l|}
    \hline
        Model & 1\% (Macro) & 5\% (Macro) & 10\% (Macro) & 25\% (Macro) & 100\% (Macro) \\ \hline
        PubMedGPT & 0.203 +/- 0.010 & 0.332 +/- 0.014 & 0.426 +/- 0.001 & 0.585 +/- 0.020 & 0.666 +/- 0.003 \\ \hline
        GatorTron & 0.296 +/- 0.006 & 0.317 +/- 0.007 & 0.407 +/- 0.015 & 0.588 +/- 0.014 & 0.677 +/- 0.008 \\ \hline
        RoBERTa & 0.388 +/- 0.014 & 0.404 +/- 0.003 & 0.520 +/- 0.043 & 0.658 +/- 0.007 & 0.690 +/- 0.010 \\ \hline
        BioClinRoBERTa & 0.310 +/- 0.004 & 0.417 +/- 0.015 & 0.524 +/- 0.018 & 0.648 +/- 0.006 & 0.707 +/- 0.007 \\ \hline
        Clinical-T5-Large & 0.356 +/- 0.007 & 0.465 +/- 0.047 & 0.548 +/- 0.012 & 0.620 +/- 0.008 & 0.663 +/- 0.007 \\ \hline
    \end{tabular}
    \caption{Macro F1 score on CLIP for models finetuned with varying amounts of annotated data. Percentages refer to fraction of the training set for the task. We report the mean and standard deviation over three random seeds. We always evaluate on the full test set.}
    \label{appendix:ablation-clip-macro}

\end{table}

\end{document}